\documentclass[10pt,twocolumn,letterpaper]{article}

\usepackage{iccv}
\usepackage{times}
\usepackage{epsfig}
\usepackage{graphicx}
\usepackage{amsmath}
\usepackage{amssymb}
\usepackage{amsmath}
\usepackage{amssymb}
\usepackage{algorithm}
\usepackage{algorithmicx}
\usepackage{algpseudocode}
\usepackage{graphics}
\usepackage{threeparttable}
\usepackage{color}
\usepackage[normalem]{ulem}
\usepackage{multirow}
\usepackage{float}
\usepackage{amsfonts}
\usepackage{bm}
\usepackage{array}
\usepackage[table]{xcolor}
\usepackage{colortbl}
\usepackage{subcaption}
\DeclareMathOperator*{\argmin}{argmin}
\usepackage[breaklinks=true,bookmarks=false]{hyperref}

\iccvfinalcopy % *** Uncomment this line for the final submission

\def\iccvPaperID{712} % *** Enter the ICCV Paper ID here

\begin{document}

%%%%%%%%% TITLE
\title{Deep Cropping via Attention Box Prediction and Aesthetics Assessment}
\author{Wenguan Wang, and Jianbing Shen\thanks{Corresponding author: Jianbing Shen (shenjianbing@bit.edu.cn). This work was supported in part by the National Basic Research Program of China (973 Program) (No. 2013CB328805), the National Natural Science Foundation of China (61272359), and the Fok Ying-Tong Education Foundation for Young Teachers. Specialized Fund for Joint Building Program of Beijing Municipal Education Commission.~}\\
 \textit{Beijing Lab of Intelligent Information Technology,} \\
 \textit{School of Computer Science, Beijing Institute of Technology, China} \\
}

\maketitle
\thispagestyle{empty}
\pagestyle{empty}

%%%%%%%%% ABSTRACT
\begin{abstract}
We model the photo cropping problem as a cascade of attention box regression and aesthetic quality classification, based on deep learning. A neural network is designed that has two branches for predicting attention bounding box and analyzing aesthetics, respectively. The predicted attention box is treated as an initial crop window where a set of cropping candidates are generated around it, without missing important information. Then, aesthetics assessment is employed to select the final crop as the one with the best aesthetic quality. With our network, cropping candidates share features within full-image convolutional feature maps, thus avoiding repeated feature computation and leading to higher computation efficiency. Via leveraging rich data for attention prediction and aesthetics assessment, the proposed method produces high-quality cropping results, even with the limited availability of training data for photo cropping. The experimental results demonstrate the competitive results and fast processing speed (5 fps with all steps).
\end{abstract}

%%%%%%%%% BODY TEXT
\section{Introduction}
\begin{figure}%%[t]
  \centering
     \includegraphics[width=0.95 \linewidth]{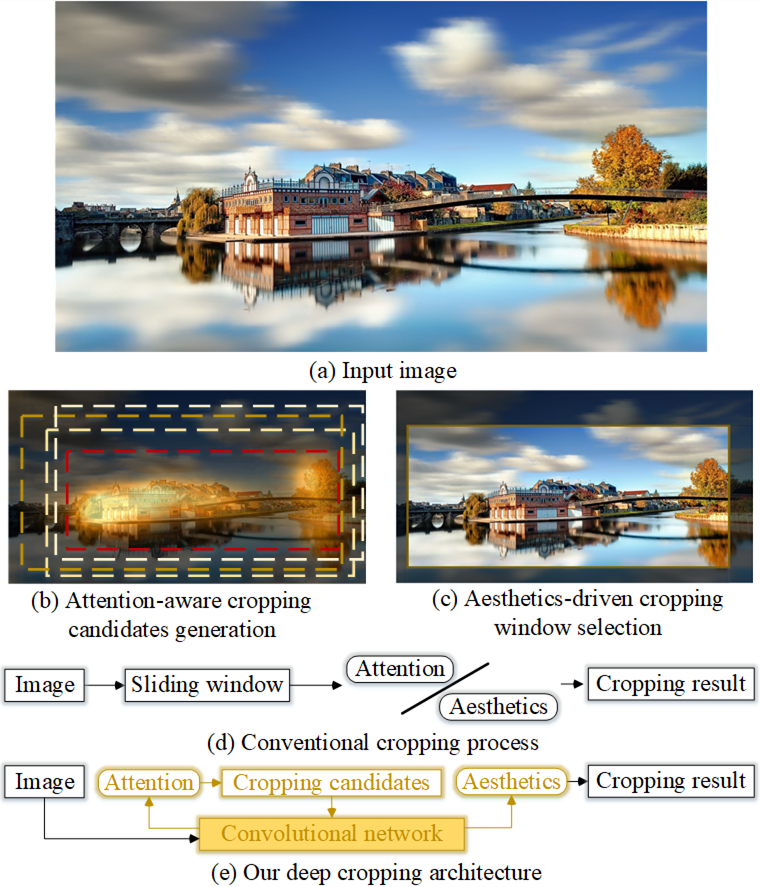}
\vspace{-4pt}
\caption{(a)-(c) Flowchart of our method. (d) Conventional methods apply sliding-judging cropping strategy, which is time-consuming and violates natural cropping procedure. (e) Our method works as a cascade of attention-aware crop candidates generation and aesthetics-based crop window selection, which handles photo cropping in a more natural
manner and is achieved by a neural network.}
 \label{fig:fig1} %%tr = 0.006, ts = 0.008
\end{figure}

Consider Fig.~\ref{fig:fig1} (a). How can we determine an appropriate crop for this picture? It seems to be a natural choice that people first define a crop that covers the desired or important region, and then, iteratively adjust the position, size and ratio of the initial crop window until achieving visual-quality-inspired result. This \textit{determining-adjusting} cropping strategy brings two advantages: (1) considering both attention and aesthetics in a cascaded way; and (2) high computation efficiency since the searching space of the best crop is only limited to the surrounding of the initial crop area. Interestingly, however, most previous cropping approaches are proceeded in another way. They usually generate a large number of sliding windows with various ratios and sizes over all the positions, and find the optimal subview via repeatedly computing attention scores \cite{marchesotti2009framework,sun2013scale,wang2016stereoscopic,Chen_2016_CVPR}, or analyzing aesthetics \cite{nishiyama2009sensation,yan2013learning} for all the sliding windows. This \textit{sliding-judging} strategy, as depicted in Fig.~\ref{fig:fig1} (d), is companied with heavy computation load, since the searching space would span all the possible subviews of the whole image. Besides, compared with repeatedly calculating attention and/or aesthetics scores over all the crop windows, arranging these two items in a sequential order would be a more reasonable and time-saving choice.

In this paper, we design a deep learning based cropping method, which models the cropping tasks as attention bounding box \textit{regression} and aesthetics \textit{classification} problems. The network is learned for directly determining the attention box that covers visually important area (the red rectangle in Fig.~\ref{fig:fig1} (b)), which seems like people first placing a crop to cover important region. Then the method generates cropping candidates (the yellow rectangles in Fig.~\ref{fig:fig1} (b)) around the attention box and selects the one with the highest aesthetics value as final crop (Fig.~\ref{fig:fig1} (c)), as the process of human iteratively adjusting initial crop and selecting the most beautiful crop window.

The proposed method approaches cropping task in a more natural and efficient way, which has the following major characteristics and contributions:\\
\noindent\textbf{Natural and unified deep cropping scheme.} The cropping procedure is arranged as a determining-adjusting process, where attention-guided cropping candidates generation is cascaded by aesthetics-aware crop window selection, as demonstrated in Fig.~\ref{fig:fig1} (e). The tasks of attention box predication and aesthetics assessment are achieved in a deep learning model, where attention information is exploited for avoiding discarding important information, while the aesthetics assessment is employed for ensuring the high aesthetic value of cropping results. The deep learning model is based on \textit{fully convolutional neural network}, which naturally supports input images of arbitrary sizes, thus avoiding undesired deformation for evaluating aesthetic quality.\\
\noindent\textbf{High computation efficiency.} Three strategies for enhancing
computational efficiency are proposed to achieve a fast processing speed of 5 fps. First, instead of searching all the possible positions in an image domain via sliding window, the approach directly regresses the attention box and generates far less number of cropping candidates ($\thicksim~\!\!\!1000$) around the visual important areas.
Second, the sub-networks of attention box prediction and aesthetics assessment share several convolutional layers in the bottom. The marginal cost for computing aesthetics estimate is decreased via sharing convolutions with attention prediction task at test-time.
Third, the approach inherits the spirit of recent object detection algorithms \cite{he2014spatial,ren2015faster,girshickICCV15fastrcnn}, which is trained to share convolutional features among cropping candidates on the feature maps. The convolutional layers are only performed once on the entire image (regardless of the number of cropping candidates), and then convolutional features of cropping candidates are extracted from feature maps, which avoiding applying the network to each cropping candidate for repeatedly computing features.\\
\noindent\textbf{Learning without sufficient cropping annotation.} For applying
deep leaning for photo cropping, an important practical catch to that solution is training data availability. The datasets for photo cropping are small-scale in deep learning terms, and primarily support evaluation. Besides, the photo cropping sometimes is a quite subjective problem which is difficult to offer a clear answer for what is a `groundtruth' crop. While the groundtruth for photo cropping is difficult to access, datasets for human gaze prediction and photo aesthetics assessment are more easily to obtain. In our method, the cropping task is explicitly achieved via learning neural network on existing rich and high-quality data for visual attention prediction and aesthetics assessment.
%as exemplified in Fig.

\section{Related Work}
\label{sec:relatedwork}
In this section, we give a brief overview of recent works in three lines: visual attention prediction, aesthetics assessment and photo cropping.
\subsection{Visual Attention Prediction}
%------------------------------------------------------------------------
Visual attention prediction aims to predict scene locations where a human observer may fixate. Early attention models \cite{itti1998model,bruce2006saliency} are typically based on various low-level features (\textit{e.g.}, color, intensity, orientation), operating and combining them at multiple scales to form a saliency map. In addition to low-level features, some approaches \cite{judd2009learning,borji2012boosting} try to employ high-level features from person or face detectors learned from specific computer vision tasks. Recently, driven by the success of deep learning in object recognition, many deep learning based attention models \cite{vig2014large,liu2015predicting,jiang2015salicon,Pan_2016_CVPR} are proposed, and generally give impressive results.
The output of traditional attention methods is usually a gray-scale image that represents the visual importance of each corresponding pixel in the image. %The attention prediction results can be used for guiding various computer vision task, such as image segmentation, visual tracking and photo cropping.
However, in our approach, we try to predict an attention bounding
box, which covers the most informative regions of the image.

\subsection{Aesthetics Assessment}
%------------------------------------------------------------------------
%manually crafting features
The main goal of aesthetics assessment is to imitate
human interpretation of the beauty of natural images. Many methods have been proposed for this topic, we refer the reader to \cite{deng2016image} for a more detailed survey.
Traditionally, aesthetic quality analysis is viewed as a
binary classification problem of predicting high- and low- quality
images. Extracting visual features and then employing various machine
learning algorithms to predict photo aesthetic values is a common pipeline in this research area.

Early methods \cite{datta2006studying,ke2006design,dhar2011high} manually designed aesthetics features
according to photographic rules or practices, such as
the rule of thirds and visual balance. Instead of using hand-crafted features, other approaches \cite{marchesotti2011assessing,su2011scenic}
have been developed to leverage more generic image descriptors, such as Fisher Vector and bag of visual words, which are previously used for image
classification but also capable of capturing aesthetic properties.
In more recent work \cite{lu2014rapid,tang2014blind,lu2015deep,kong2016photo,mai2016composition}, deep learning methods have been used to aesthetics assessment and have shown promising results.

\subsection{Photo Cropping}
%------------------------------------------------------------------------
Cropping is an important operation for improving visual quality of digital photos, which cuts away unwanted areas outside of
a selected rectangular region.  A lot of methods have been proposed towards automating this task. These methods, in general, can be categorized into \textit{attention-based} or \textit{aesthetics-based} approaches.
The attention-based approaches \cite{marchesotti2009framework,sun2013scale,Chen_2016_CVPR} focus on preserving
the main subject or visually important area in the scene after cropping. These methods usually place the crop window over
the most visually significant regions according to certain attention scores \cite{Wang2015,wang2015consistent,wang2016correspondence,Wang2017}.
The other major direction of cropping methods is
aesthetics-based approach that emphasizes the general attractiveness of the cropped image. Those aesthetics-based approaches \cite{nishiyama2009sensation,yan2013learning} are centered on composition-related image properties. Taking various aesthetical factors into account, they try to find the cropping candidate with the highest quality score.

In this paper, we consider both attention and aesthetics information, which are arranged in a natural and cascaded manner. The proposed method approaches photo cropping as a cascade of generating cropping candidates via attention box prediction and selecting best crop according to aesthetics criteria. Our method shares the spirit of recent object detection algorithms \cite{he2014spatial,ren2015faster,girshickICCV15fastrcnn}, one branch of our network learns to predict the bounding box covers visually important area, while the other one tries to analyze aesthetic value.

\section{Our Approach}
\label{sec:3}
%\subsection{Algorithm Overview}
%\label{sec:3.1}
%-------------------------------------------------------------------------
%Our method approaches photo cropping in a determining-adjusting manner, which infers initial crop as a bounding box covering the visually important area (attention-aware determining), and then selects the best crop with highest aesthetic quality from a few crop candidates generated around the initial crop (aesthetic-based adjusting). Two key subproblems are abstracted from above process: (1) attention box prediction for determining the initial crop; (2) aesthetics assessment for determining the final crop. For achieving above subtasks, we design a deep learning model that has two submodule networks: Attention Box Prediction (ABP) network and Aesthetics Assessment (AA) network.

The cropping algorithm is decomposed into two cascaded stages, namely, attention-aware cropping candidates generation (Sec.~\ref{sec:3.1}) and aesthetics-based crop window selection (Sec.~\ref{sec:3.2}). It infers initial crop as a bounding box covering the most visually important area, and then selects the best crop with highest aesthetic quality from a few crop candidates generated around the initial crop. We design a deep learning model that has two sub-networks: Attention Box Prediction (ABP) network and Aesthetics Assessment (AA) network, for achieving two key subtasks in above cropping process: (1) attention box prediction for determining the initial crop; and (2) aesthetics assessment for determining the final crop. Those two networks share several convolutional blocks in the bottom and are based on fully convolutional network, which will be detailed in following sections. Finally, in Sec.~\ref{sec:3.3}, we will give more details of our model in training and testing.

%\begin{figure}%%[t]
%  \centering
%     \includegraphics[width=0.99 \linewidth]{figs/fig2.png}
%\caption{Architecture of our deep cropping model. It consists of two sub-networks: Attention Box Prediction (ABP) network and Aesthetics Assessment (AA) network, which share several convolution layers at the bottom.}
% \label{fig:fig2} %%tr = 0.006, ts = 0.008
%\end{figure}
%approach is guided by two principles: (1) from the view of cropping task, the method is proceeded in a determining-adjusting manner; and (2) from the perspective of neural network, deep cropping is achieved by a deep learning model which simultaneously learns two submodules for attention box prediction and aesthetics assessment.

\subsection{Attention-aware Cropping Candidates}
\label{sec:3.1}
%-------------------------------------------------------------------------
In this section, we introduce our method for cropping candidates generation, which is based on an Attention Box Prediction (ABP) network. This network takes an image of any size as input and outputs a set of rectangular crop windows, each with a score that stands for the prediction accuracy. Then the initial crop is identified as the most accurate one, and various cropping candidates with different sizes and ratios are generated around it. After that, the final crop is selected from those candidates according to their aesthetic quality based on an Aesthetics Assessment (AA) network (Sec.~\ref{sec:3.2}).

The initial crop can be viewed as a rectangle that preserves the most informative part of the image while has minimum area. This optimal rectangle searching problem is a common task for attention-based cropping methods. Let $P\in [0, 1]^{w\times h}$ be an attention mask, we first define a set of crop windows $\mathfrak{W}$:
\begin{equation}
    \begin{aligned}
    \mathfrak{W} = \{W|\sum_{x \in W}P(x) > \lambda \sum_{x}P(x)\},
    \end{aligned}
    \label{eq:1}
\end{equation}
where $\lambda \in [0,1]$ is a fraction threshold. Then the optimum rectangle $\dot{W}$ is defined as:

\begin{equation}
    \begin{aligned}
    \dot{W} = \mathop{\argmin}_{W \in \mathfrak{W}}|W|.
    \end{aligned}
    \label{eq:2}
\end{equation}

Equ. \ref{eq:2} can be solved via sliding window with $\mathcal{O}(w^2h^2)$ computation complexity, while a recent method \cite{Chen_2016_CVPR} shows it can be solved with computation complexity of $\mathcal{O}(wh^2)$.
\begin{figure}%%[t]
  \centering
     \includegraphics[width=0.99 \linewidth]{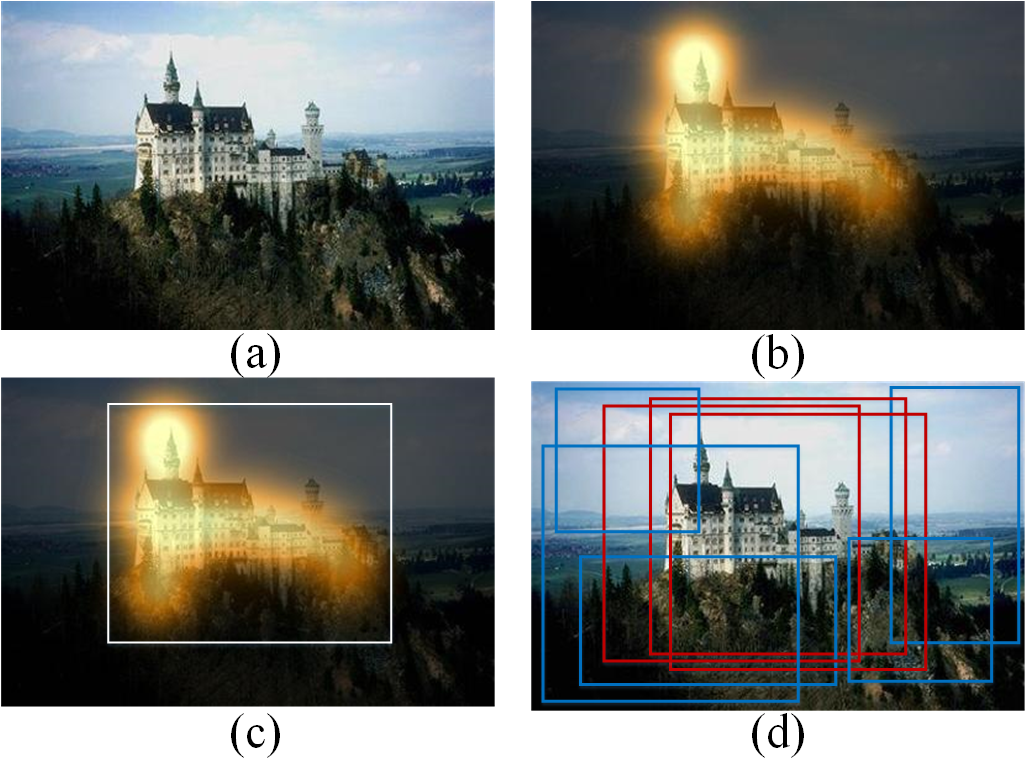}
\caption{(a) Input image. (b) Attention map. (c) Ground truth attention box generation via \cite{Chen_2016_CVPR}. (d) Positive (red) and negative (blue) defaults boxes are generated for training ABP network according to ground truth attention box.}
 \label{fig:fig3} %%tr = 0.006, ts = 0.008
\end{figure}

Differently, we design a neural network for directly predicting such attention box. Given a training sample $(I, G)$ consisting
of an image $I$ of size $w\times h \times 3$ (Fig.~\ref{fig:fig3} (a)), and a groundtruth attention map $G\in [0, 1]^{w\times h}$ (Fig.~\ref{fig:fig3} (b)), the optimum rectangular $\dot{W}$ defined in Equ. \ref{eq:2} is computed as the groundtruth attention prediction box. Here we apply \cite{Chen_2016_CVPR} for generating $\dot{W}$ over $G$ (Fig.~\ref{fig:fig3} (c)) for computation efficiency. We set $\lambda = 0.9$ for preserving most informative areas. Then the task of attention box prediction can be achieved via bounding box regression as object detection \cite{he2014spatial,ren2015faster,girshickICCV15fastrcnn}. Note that any other attention scores can also be used for generating groundtruth bounding box for training the ABP network.

\begin{figure}%%[t]
  \centering
     \includegraphics[width=0.99 \linewidth]{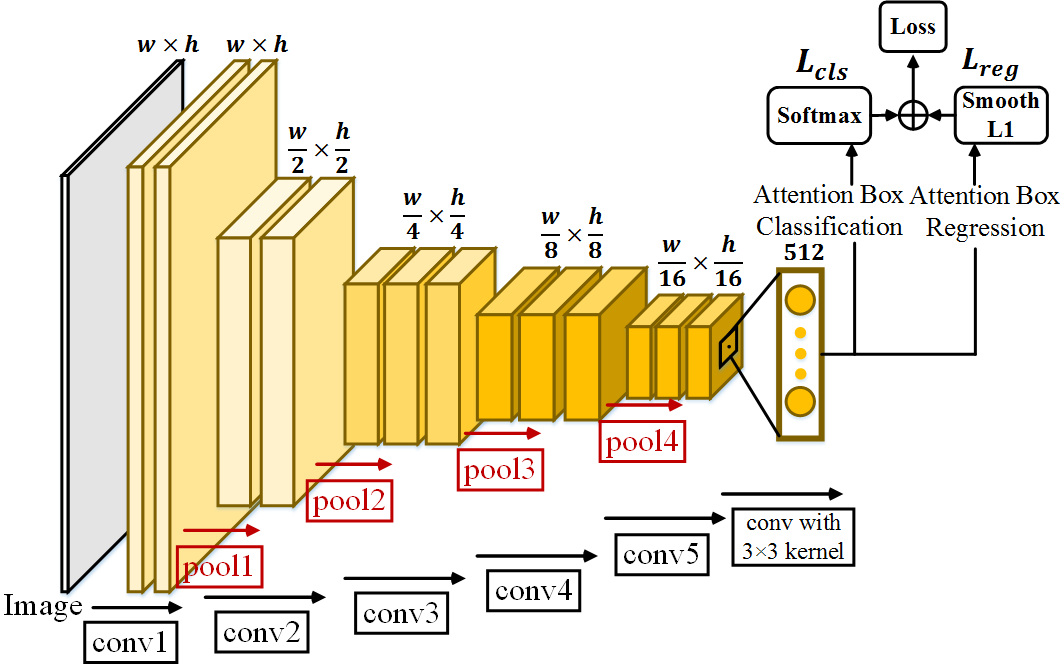}
\caption{Architecture of Attention Box Prediction (ABP) network.}
 \label{fig:fig4} %%tr = 0.006, ts = 0.008
\end{figure}

Fig.~\ref{fig:fig4} illustrates the architecture of ABP network. The bottom of this network is a stack of convolutional layers, which are borrowed from the first five convolutional blocks of VGGNet \cite{simonyan2014very}. With the last convolutional layer, we slide a small network with $3 \times 3$ kernel over its convolutional feature map, thus generating $512\!-\!d$ feature for each sliding location. The feature vector is further fed into two fully-connected layers: box-regression layer for predicting attention bounding box; box-classification layer for determining the box whether belongs to attention box. For a given location, those two fully-connected layers predict box offsets and scores over a set of default bounding boxes, which are similar to the \textit{anchor boxes} used in Faster R-CNN \cite{ren2015faster}.

During training, we need to determine which default boxes correspond to the groundtruth attention box and train the network accordingly. We assign the default box which has the highest Intersection-over-Union (IoU) with the groundtruth box or with IoU higher than 0.7 as a positive label ($c=\mathbf{1}$). We assign the
default box that has a IoU lower than 0.3 a negative label ($c=\mathbf{0}$) and drop other default boxes. The above process is illustrated in Fig.~\ref{fig:fig3} (d).
For the preserved boxes, we define $\bar{p}^c_i \in \{1,0\}$ as an indicator for the label of $i$-th
box and vector $\bar{t}$ as a four-parameterized coordinate (coordinates of center, width and height) of the groundtruth attention box. Similarly, we define $p^c_i$ and $t_i$ as predicted confidence over $c$ class and predicted attention box of $i$-th default box.
With above definition, the ABP network is trained via minimizing the following loss function derived from object detection \cite{girshick2014rich,ren2015faster,liu2016ssd}:
\begin{equation}
    \begin{aligned}
    \mathcal{L}(p, t) = \sum\nolimits_{i}\mathcal{L}_{cls}(p_i,\bar{p}_i)+\sum\nolimits_{i}\bar{p}^\mathbf{1}_i\mathcal{L}_{reg}(t_i,\bar{t}).
    \end{aligned}
    \label{eq:3}
\end{equation}
%\begin{equation}
%    \begin{aligned}
%    \mathcal{L}(\!\{p_i\}\!,\!\{t_i\}\!)\!\! = \!\! \mathcal{L}_{cls}(\{p_i\},\{p^*_i\})\!+\!\lambda\!\sum_{i}p^*_i\mathcal{L}_{reg}(t_i,t^*_i).
%    \end{aligned}
%    \label{eq:3}
%\end{equation}
%Here, $p_i$ indicates the predicated probability of the $i$-th default box being the attention box. $p^*_i$ is the ground truth label, which is 1 if the default box is positive, and is 0 if it is negative. $t_i$ is a four-parameterized coordinate vector (coordinates of center, width and height) of the predicted bounding box, and $t^*_i$ is that of the ground-truth box associated with a positive default box.
The classification loss $\mathcal{L}_{cls}$  is the softmax loss over confidences of two classes (attention box or not). The regression loss $\mathcal{L}_{reg}$ is a Smooth L1 loss \cite{girshick2014rich}, between the predicated box and the ground truth attention box, which is only activated for positive default boxes. %The weight term $\lambda$ is set to 1 according to \cite{ren2015faster,liu2016ssd}.

With the ABP network trained on existing attention prediction datasets, it learns to generates reliable attention boxes. Then we select the one with the highest prediction score (${p}^\mathbf{1}_i$) as the initial crop. This initial crop covers the most informative part of the image, which likes human placing a crop around the desired area (Fig.~\ref{fig:fig5} (a)). Next, we generate a set of cropping candidates around the initial crop, as the human adjusting the location, size and ratio of the initial crop. A rectangular can be uniquely determined via the coordinates of its top-left and right-bottom corners. For the top-left corner of the initial crop, we define a set of offsets: $\{-40,-32,\cdots,-8,0\}$ in x- and y-axis. Similarly, a set of offsets: $\{0,8,...,32,40\}$ in x- and y-axis is also defined for the bottom-right corner.
Via adding the top-left and bottom-right corners with corresponding pre-defined offsets \footnote{Since we resize the input image with $min(w,h) = 224$, we find the largest offset (40) is enough.}, we generate $6^4 = 1296$ cropping candidates in total, which is far less than the sliding windows needed for traditional cropping methods. Each of crop candidates is designed for covering the whole initial crop area, since the initial crop is a minimum visually importance-preserved rectangle that should be maintained in cropping process (Fig.~\ref{fig:fig5} (b)).
%-------------------------------------------------------------------------

\subsection{Aesthetics-based Crop Window Selection}
\label{sec:3.2}
%-------------------------------------------------------------------------
\begin{figure}%%[t]
  \centering
     \includegraphics[width=0.99 \linewidth]{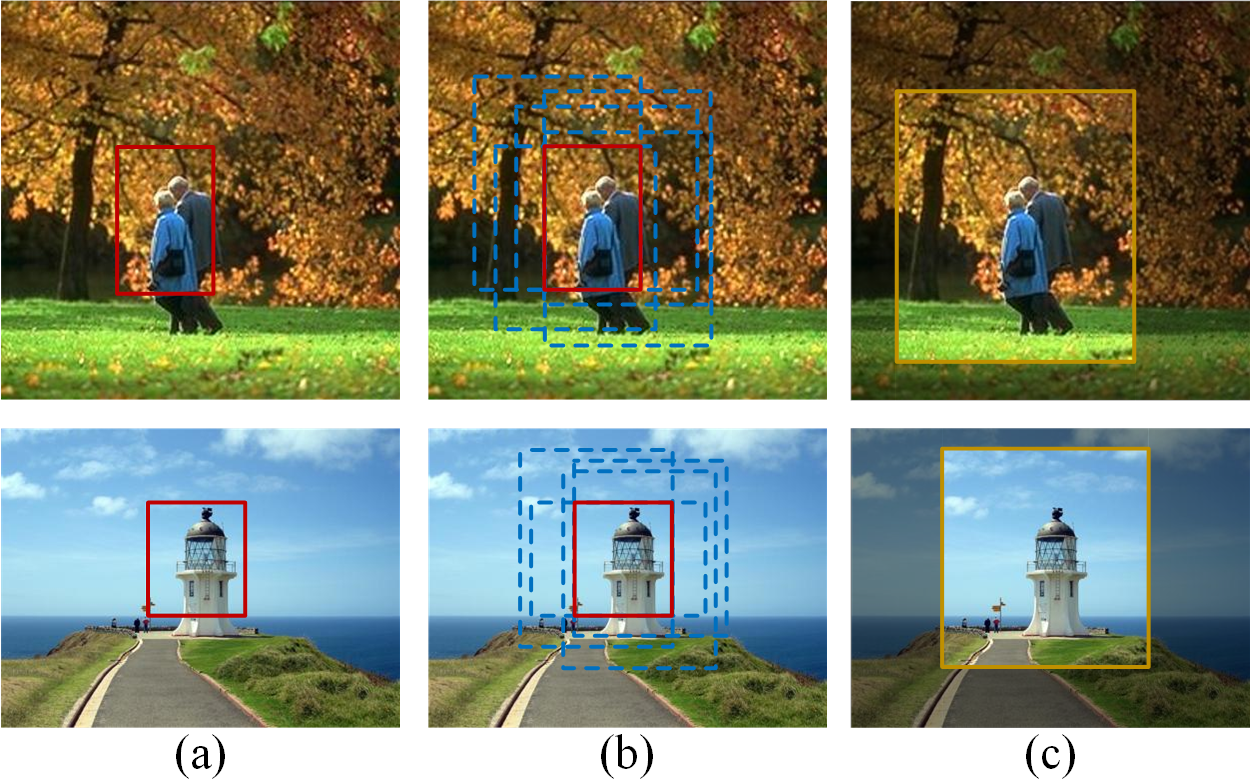}
\caption{(a) Initial crop (red rectangle) predicted via ABP network. (b) Cropping candidates (blue rectangles) generated around the initial crop. (c) Final crop selected as the candidate with highest aesthetic score from AA network.}
 \label{fig:fig5} %%tr = 0.006, ts = 0.008
\end{figure}
With our attention-aware cropping candidates by ABP network, we next select the most aesthetics-inspired one as the final crop. It is important to consider aesthetics for photo cropping task, since beyond preserving the important content, a nice crop should also deliver pleasant viewing experience.
For analyzing the aesthetic quality of each cropping candidates, one choice is training an aesthetics assessment network, and iteratively applying forward-propagation for each crop candidate over this network when cropping. Obviously, this strategy is straightforward but time-consuming. Inspired by the recent advantages of object detection, which share convolutional features between regions, we build a network that analyzes aesthetic values of all cropping candidates simultaneously.

\begin{figure}%%[t]
  \centering
     \includegraphics[width=0.99 \linewidth]{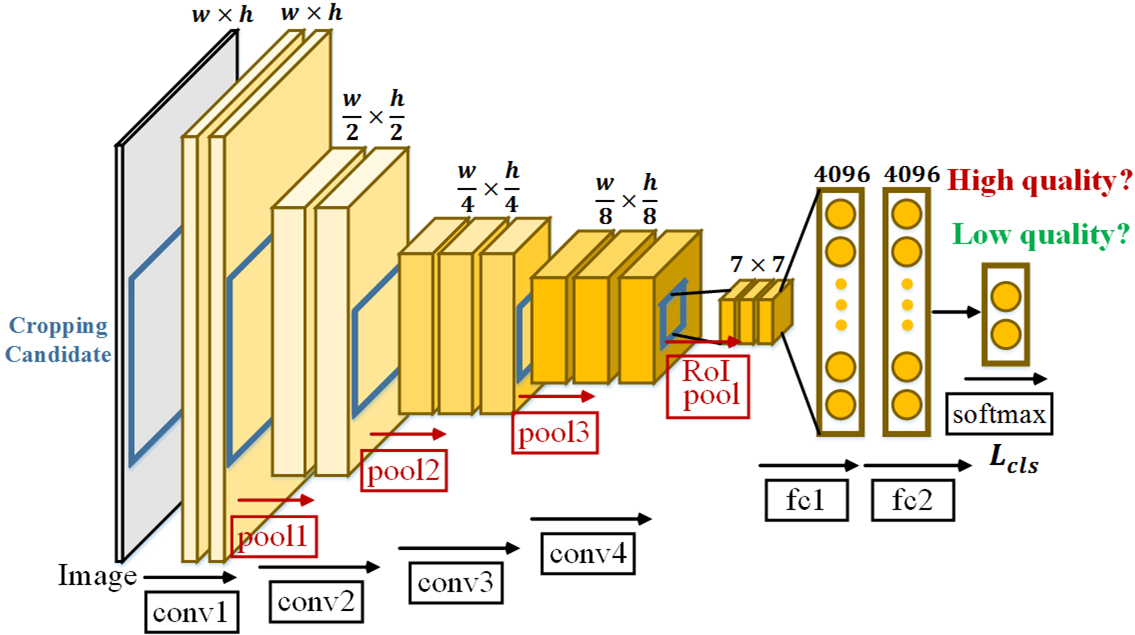}
\caption{\small{Architecture of Aesthetics Assessment (AA) network.}}
 \label{fig:fig6} %%tr = 0.006, ts = 0.008
\end{figure}

We achieve this via an Aesthetics Assessment (AA) network (Fig.~\ref{fig:fig6}), which takes an entire image and a set of cropping candidates as input, and outputs the aesthetic values of the cropping candidates. The bottom of the AA network is the former four convolutional blocks of VGGNet \cite{simonyan2014very} without $pool4$ layer. Here we adopt a relatively shallow network mainly due to two reasons. First, aesthetics assessment is a relatively easier problem (high quality \textit{vs} low quality) compared with image classification (with 1000 classes). Secondly, for an image with the size of $w\times h\times 3$, the spatial dimensions of the final convolutional feature map of AA network is $\frac{w}{8}\times \frac{h}{8}$, which preserves discriminability for the offsets defined in Sec.~\ref{sec:3.1}.

Then, on the top of the last convolutional layer, we adopt Region of Interest (RoI) pooling layer  \cite{ren2015faster}, which is a special case of spatial pyramid pooling (SPP) layer \cite{he2014spatial}, to extract a fixed-length feature vector from the final convolutional feature map. The RoI pooling layer uses max pooling to convert the features inside any crop candidate into a small feature map with a fixed-dimensional vector, which is further fed into a sequence of fully-connected layer for aesthetic quality classification. This operation allows us to operate image with arbitrary aspect ratios, thus avoiding undesired deformation in aesthetics assessment. With a crop candidate with size of $w'\times h'$, RoI pooling layer divides it into $n\times n$ spatial bins and applies max-pooling for the features within each bins. Here we set $n = 7$.

For training, given an image from the existing aesthetics assessment datasets, it takes an aesthetic label $c\in \{\mathbf{1},\mathbf{0}\}$, where $\mathbf{1}$ corresponds to high aesthetic quality and $\mathbf{0}$ represents low quality. We resize the image with $min(w,h) = 224$, similar to ABP net, and the whole image can be viewed as a cropping candidate for training. For $i$-th image in training, we define $\bar{q}^c_i \in \{1,0\}$  as an indicator for its aesthetics-quality label and ${q}^c_i$ is its predicted aesthetics-quality score for $c$ class.

Based on the above definition, the training of the AA network is done by minimizing the following softmax loss over $N$ training samples:
\begin{equation}
    \begin{aligned}
    &\mathcal{L}_{cls}(q,\bar{q}) = -\frac{1}{N}\sum_i\sum_{c \in \{\mathbf{1},\mathbf{0}\}} \bar{q}^c_i log (\widehat{q}^c_i),\\
    &where ~~ \widehat{q}^c_i = exp(q^c_i)\Big/(exp(q^\mathbf{1}_i)+exp(q^\mathbf{0}_i)).
    \end{aligned}
    \label{eq:4}
\end{equation}

With the cropping candidates generated from APB network, the AA network is capable of producing their aesthetics-quality scores ($\{{q}^\mathbf{1}_i\}$), where the one with the highest score is selected as the final crop (Fig.~\ref{fig:fig5} (c)).
\begin{figure*}%%[t]
  \centering
        \includegraphics[width = 0.99 \linewidth]{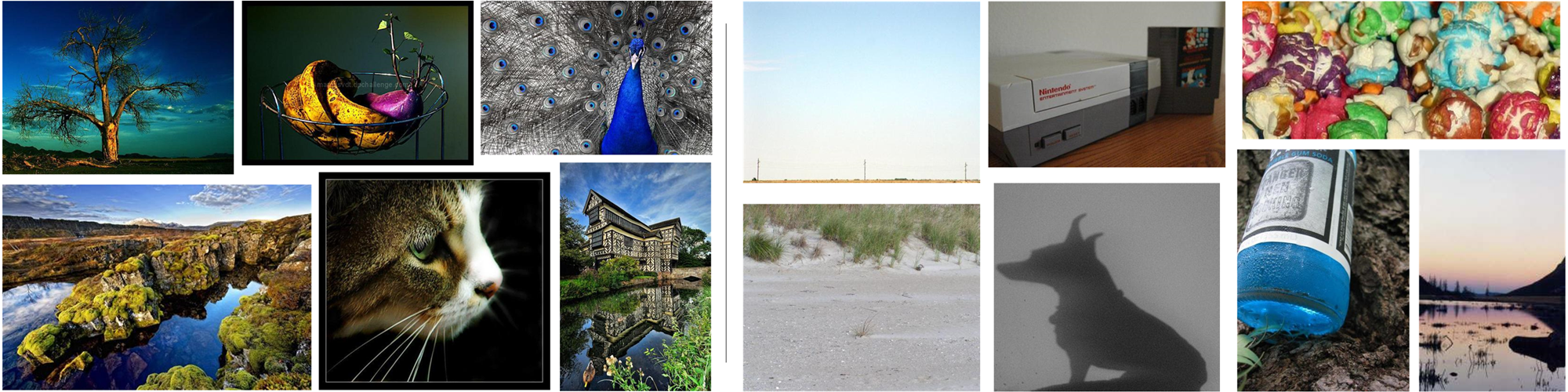}\\
  \hfill\mbox{}(a) Images with highest aesthetics values\hfill\mbox{}
  \hfill\mbox{}(b) Images with lowest aesthetics values\hfill\mbox{}
\caption{Aesthetics assessment results via our AA network. The test images with the highest predicted aesthetics values
and those with the lowest predicted aesthetics values are presented in (a) and (b), respectively.}
 \label{fig:fig8} %%tr = 0.006, ts = 0.008
\end{figure*}
\subsection{Implementation Details}
\label{sec:3.3}
\noindent\textbf{Training} Two large-scale datasets: SALICON \cite{jiang2015salicon} and AVA \cite{murray2012ava}, are used for training our model.
SALICON is used for training our ABP network. It contains 15000 natural images with eye fixation annotations which are simulated through mouse movements of users on blurred images. For obtaining groundtruth attention box, we follow the instructions of \cite{jiang2015salicon} for transferring the binary mouse-clicking map into grey-scale human attention map, and then we apply \cite{Chen_2016_CVPR} for generating attention bounding box according to Equ. \ref{eq:2} with $\lambda = 0.9$. The AVA dataset is the largest publicly available aesthetics assessment benchmark, which provides about 250,000 images in total. The aesthetics quality of each image was rated on average by roughly 200 people with the ratings ranging from one to ten, with ten indicating the highest aesthetics quality. Followed by \cite{lu2014rapid,lu2015deep,mai2016composition,murray2012ava}, about 230,000 images are used for training our AA network. More specially, images with mean ratings smaller than 5 are assigned as low quality and those with mean ratings larger than or equal to 5 are labeled as high quality.

Our two sub-networks are trained simultaneously. In each training iteration, we use a min-batch of 4 images, 2 of which are from SALICON dataset with the groundtruth attention boxes and the rest from AVA dataset with aesthetics quality groundtruth. Before feeding the input images and ground-truth to the network, we scale the images such that the smaller dimension is 224. Since the bottom two convolutional blocks ($conv1$ and $conv2$) are shared between both the tasks of attention box prediction and esthetics assessment, they are trained for the two tasks simultaneously using all the images in the batch. For the layers specialized to each of the sub-networks are trained using only those images in the batch having the corresponding ground-truth.

Both ABP and AA networks are initialized from the weights of VGGNet \cite{simonyan2014very}, which is pre-trained on large-scale image classification dataset \cite{russakovsky2015imagenet}. Our model is implemented with the popular
Caffe library \cite{jia2014caffe} and trained with stochastic gradient descent. The networks were trained over 200K
iterations where we use momentum of 0.9 and weight decay of 0.0001, which is reduced by
a factor of 0.1 at every 10K iterations.

\noindent\textbf{Testing} For training, our two sub-networks are trained in parallel strategy, while for testing, they work in a cascaded way. With a given image (resized with $min(w,h) = 224$) for cropping, we first gain a set of attention boxes generated via forward propagation on APB network. Then the initial crop was selected as the one with the highest score of attention box prediction. After that, a set of cropping candidates are generated around the initial crop. Since the two convolutional blocks at the bottom are shared between ABP and AA networks, we directly feed the cropping candidates and the convolutional feature of last layer of $conv2$ into AA network. Finally, the final crop is selected as the cropping candidate with best aesthetic quality. The cropping model achieves a fast speed of 5 fps.

\section{Experimental results}
\label{sec:4}
In this section, we first examine the performance of our ABP and AA networks on their specific tasks. The goal of these experiments is to investigate the effectiveness of individual components instead of comparing them with the state-of-the-art. Then, we evaluate the performance of our whole cropping model on two widely used photo cropping datasets with other competitors.
%-------------------------------------------------------------------------
\subsection{Evaluation for ABP and AA Networks}
%-------------------------------------------------------------------------
%Although the attention box prediction and aesthetics assessment are not our main goal, we still examine their performance since they are key components of the proposed cropping model.
\noindent\textbf{Performance of ABP Network} We first evaluate the performance of ABP network on PASCAL dataset \cite{li2014secrets}, which is widely used for attention prediction. This dataset contains totally 850 natural images from PASCAL 2010 \cite{everingham2010pascal}, with the eye fixations during 2 seconds of 8 different subjects. With the binary eye fixation images, we follow \cite{li2014secrets} to generate gray-scale attention map. Then, as the way described in Sec. \ref{sec:3.3}, we generate groundtruth attention box for each image. We consider eight state-of-the-art attention models: ITTI \cite{itti1998model}, AIM \cite{bruce2006saliency}, GBVS \cite{harel2006graph}, SUN \cite{zhang2008sun}, DVA \cite{hou2009dynamic}, SIG \cite{hou2012image}, CAS \cite{goferman2012context} and SalNet \cite{Pan_2016_CVPR}. Then we extract the attention boxes of above methods via the same strategy used for generating groundtruth bounding box. We opt for the Intersection over Union (IoU) score for quantifying the quality of extracted attention boxes. The quantitative results are illustrated in Table \ref{table1}. As seen, our attention box prediction results are more accurate
than previous attention models, since our ABP network is specially designed for this task.
\begin{table}[!htbp]
\centering
\resizebox{0.49\textwidth}{!}{
\begin{tabular}{|c||c|c|c|c|c|}
\hline
\!\!\!Method\!\!\! &\textbf{Ours} &ITTI\cite{itti1998model} &AIM \cite{bruce2006saliency} &GBVS\cite{harel2006graph}&SUN\cite{zhang2008sun} \\
 \hline
IoU    &\textbf{0.517} &0.318 &0.327 & 0.319& 0.273 \\
\hline
 \hline
\!\!\!Method\!\!\! &\textbf{Ours} &DVA\cite{hou2009dynamic} &SIG\cite{hou2012image} &CAS \cite{goferman2012context}&SalNet \cite{Pan_2016_CVPR}\\
\hline
IoU    &\textbf{0.517} &0.346  &0.272 &0.356 &0.379\\
\hline
\end{tabular}
}
\caption{\small{Attention box prediction with IoU for PASCAL \cite{li2014secrets}.}}
\label{table1}
\end{table}

\noindent\textbf{Performance of AA Network} We adopt the testing set of AVA dataset \cite{murray2012ava}, which is mentioned in Sec. \ref{sec:3.3}, for evaluating the performance of our AA network. The testing set of AVA dataset contains 19,930 images. The testing images with mean ratings smaller than 5 are labeled as low quality; otherwise they are labeled as high quality. We compare our methods with the state-of-the-art methods: AVA \cite{murray2012ava}, RAP \cite{lu2014rapid}, RAP2 \cite{lu2015R},  DMA \cite{lu2015deep},  ARC \cite{kong2016photo} and CPD \cite{mai2016composition}, where AVA is based on manually designed features while other methods are based on deep learning model. As shown in Table \ref{table2}, our AA network is struggle to achieve state-of-the-art performance due to relatively simple network architecture. In Fig.~\ref{fig:fig8}, we present some examples of
the test images that are considered of the highest and lowest aesthetics values by our AA network.

\begin{table}
\centering
\resizebox{0.49\textwidth}{!}{
\begin{tabular}{|c||c|c|c|c|}%{|p{1cm}|p{1cm}|p{1cm}|p{2cm}|}  % {lccc}
\hline
Method &\!\!\textbf{Ours}\!\! &AVA~\!\cite{murray2012ava} &\!\!\!\!RAP-DCNN~\!\cite{lu2014rapid}\!\!\!\!\!\! &\!\!\!RAP-RDCNN~\!\cite{lu2014rapid}\!\!\!\! \\
 \hline
\!\!Accuracy\!\!    &\!\!\textbf{0.769}\!\! &0.667 &0.732 &0.745 \\
\hline
 \hline
Method &\!\!\textbf{Ours}\!\! &RAP2~\!\cite{lu2015R} &DMA-SPP~\!\cite{lu2015deep}&DMA~\!\cite{lu2015deep}\\
 \hline
\!\!Accuracy\!\! &\!\!\textbf{0.769}\!\!&0.754  &0.728   &0.745 \\
\hline
 \hline
Method &\!\!\textbf{Ours}\!\! &\!\!\!DMA-Alex~\!\cite{lu2015deep}\!\!\!\!   &ARC~\!\cite{kong2016photo}&CPD~\!\cite{mai2016composition}\\
 \hline
\!\!Accuracy\!\!    &\!\!\textbf{0.769}\!\! &0.754  &0.773 &0.774\\
\hline
\end{tabular}
}
\caption{Aesthetics assessment accuracy for AVA \cite{murray2012ava}.}
\label{table2}
\end{table}

\noindent\textbf{Conclusion} Overall, our two sub-networks generate the promising results or compete with existing top-performance approaches. Considering the shared convolutional layers in the bottom of these two networks, our model achieves a good tradeoff between performance and computation efficiency. More important, the robustness of those two basic components greatly contributes the high-quality of our crop suggestions, which will be detailed in next section.

\subsection{Evaluation for Photo Cropping}
We evaluate our whole cropping model on two public image cropping datasets, including Image Cropping Dataset from MSR (MSR-ICD) \cite{yan2013learning} and FLMS \cite{fang2014automatic}. The MSR-ICD dataset includes 950 images and each image is carefully cropped by 3 experts. The FLMS dataset contains 500 natural images which are collected from Flickr. For each image, 10 expert users on Amazon Mechanical Turk who passed a strict qualification test are employed for cropping groundtruth box.

We adopt the same evaluation metrics as \cite{yan2013learning}, \textit{i,e.}, IoU score and Boundary Displacement Error (BDE), to measure the cropping accuracy of image croppers. BDE is defined as the mean normalized displacement of four edges between the cropping box and the groundtruth rectangles.

\begin{table}[!htbp]
\centering
\resizebox{0.49\textwidth}{!}{
\begin{tabular}{|c||c|c||c|c||c|c|}  % {lccc}
\hline
\multirow{2}*{\!\!\!\!Method\!\!\!\!}
&\multicolumn{2}{c||}{\!\!\!Photographer \!1\!\!\!} & \multicolumn{2}{c||}{\!\!\!Photographer \!2\!\!\!}  & \multicolumn{2}{c|}{\!\!\!Photographer \!3\!\!\!}\\
\cline{2-7}
&\!\!IoU$\uparrow$\!\!&\!\!BDE$\downarrow$\!\!&\!\!IoU$\uparrow$\!\!&\!\!BDE$\downarrow$\!\!&\!\!IoU$\uparrow$\!\! &\!\!BDE$\downarrow$\!\!\\
\hline
\hline
\!\!\!ATC \cite{suh2003automatic}\!\!\! &\!\!0.605\!\!&\!\!0.108\!\! 	&\!\!0.628\!\! 	&\!\!0.100\!\! 	&\!\!0.641\!\! 	&\!\!0.095\!\!\\
\!\!\!AIC \cite{Chen_2016_CVPR}\!\!\!&\!\!0.469\!\!&\!\!0.142\!\! 	&\!\!0.494\!\! 	&\!\!0.131\!\! 	&\!\!0.512\!\! 	&\!\!0.123\!\!\\
\!\!\!LCC \cite{yan2013learning}\!\!\!&\!\!0.748\!\!&\!\!0.066\!\! 	&\!\!0.728\!\! 	&\!\!0.072\!\! 	&\!\!0.732\!\! 	&\!\!0.071\!\!\\
\!\!\!MPC \cite{park2012modeling}\!\!\!&\!\!0.603\!\!&\!\!0.106\!\! 	&\!\!0.582\!\! 	&\!\!0.112\!\! 	&\!\!0.608\!\! 	&\!\!0.110\!\!\\
\!\!\!SPC \cite{nishiyama2009sensation}\!\!\!&\!\!0.396\!\!&\!\!0.177\!\! 	&\!\!0.394\!\! 	&\!\!0.178\!\! 	&\!\!0.385\!\! 	&\!\!0.182\!\!\\
\!\!\!ARC \cite{kong2016photo}\!\!\!&\!\!0.448\!\!&\!\!0.163\!\! 	&\!\!0.437\!\! 	&\!\!0.168\!\! 	&\!\!0.440\!\! 	&\!\!0.165\!\!\\
\textbf{Ours}&\!\!\textbf{0.813}\!\! &\!\!\textbf{0.030}\!\!&\!\!\textbf{0.806}\!\! &\!\!\textbf{0.032}\!\!	&\!\!\textbf{0.816}\!\! 	&\!\!\textbf{0.032}\!\!\\
\hline
\end{tabular}
}
\label{table3}
\caption{\small{Cropping results with IoU and BDE on MSR-ICD \cite{yan2013learning}.}}
\label{table3}
\end{table}

We compare our cropping method with two main categories of image cropping methods, \textit{i.e.}, \textit{attention-based} and \textit{aesthetics-based} methods. For attention-based method, we select ATC \cite{suh2003automatic} which is a classical image thumbnail cropping method.
We also use AIC as a baseline, which is obtained via equipping crop window researching method \cite{Chen_2016_CVPR} with top-performing saliency detection method.
\begin{table}[!htbp]
\centering
\resizebox{0.49\textwidth}{!}{
\begin{tabular}{|c||c|c|c|c|c|c|}%{|p{1cm}|p{1cm}|p{1cm}|p{2cm}|}  % {lccc}
\hline
\!\!\!Method\!\!\! &\!\!\!\textbf{Ours}\!\!\! &\!\!\!ATC \cite{suh2003automatic}\!\!\! &\!\!\!AIC \cite{Chen_2016_CVPR}\!\!\!&\!\!\!LCC \cite{yan2013learning}\!\!\!&\!\!\!MPC \cite{park2012modeling}\!\!\!&\!\!\!VBC \cite{fang2014automatic}\!\!\!\\
 \hline
IoU$\uparrow$    &\textbf{0.81} &0.72 &0.64 &0.63 &0.41 & 0.74  \\
BDE$\downarrow$    &\textbf{0.057} &0.063 &0.075 &- &- & -  \\
\hline
\end{tabular}
}
\caption{Cropping results with IoU and BDE on FLMS \cite{fang2014automatic}.}
\label{table4}
\end{table}
\begin{figure}[!htbp]
  \centering
     \includegraphics[width=0.99 \linewidth]{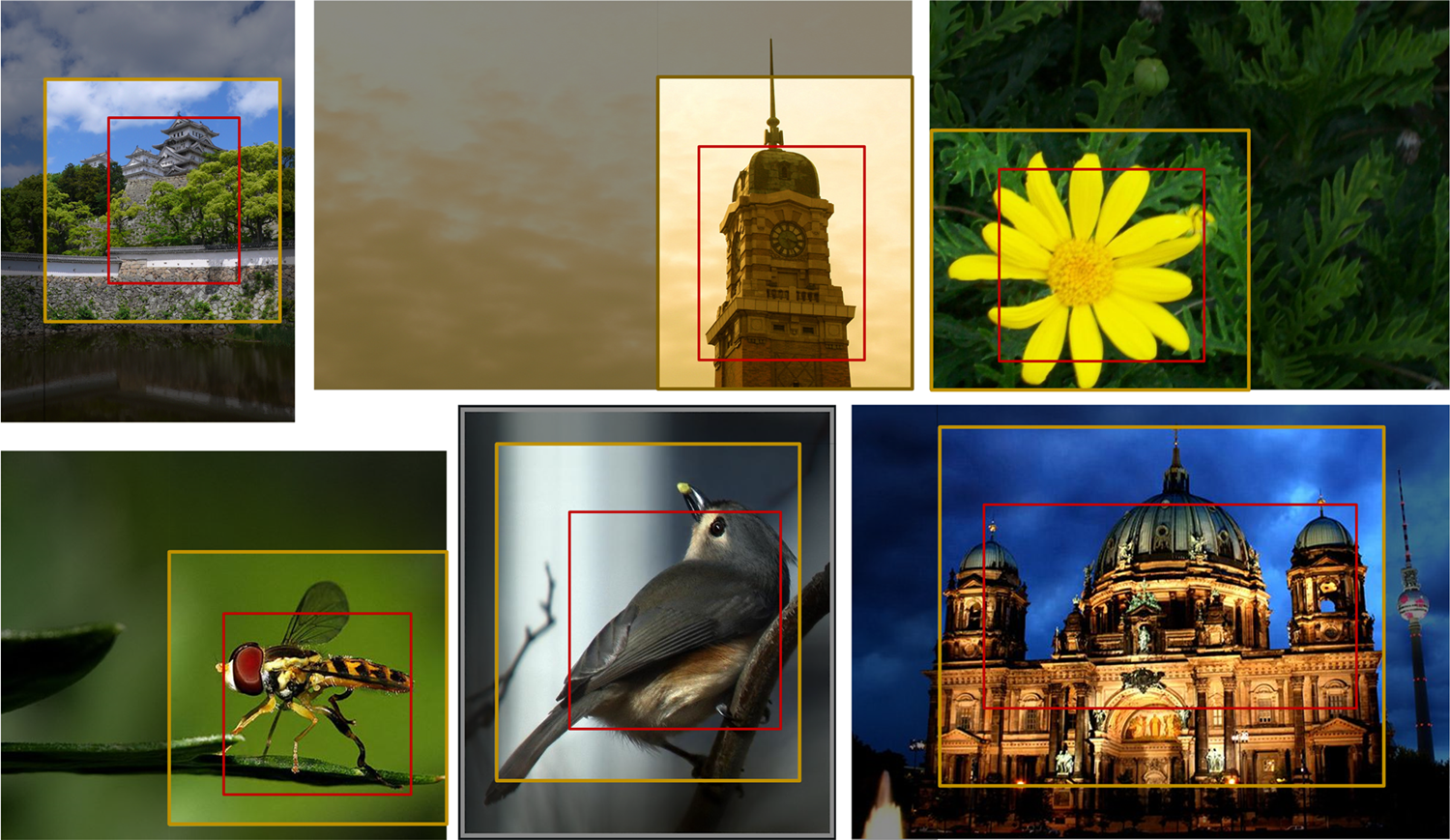}
\caption{Qualitative results on MSR-ICD \cite{yan2013learning} and FLMS \cite{fang2014automatic} datasets. The red rectangles indicate the initial crop generated via ABP network, and the yellow windows correspond to the final crop selected via AA network.}
 \label{fig:fig10} %%tr = 0.006, ts = 0.008
 \vspace{-4mm}
\end{figure}
We apply context-aware saliency \cite{goferman2012context} and optimal parameters, as suggested by \cite{Chen_2016_CVPR}, for maximizing its performance. For aesthetics-based method, we select LCC \cite{yan2013learning}, MPC \cite{park2012modeling}, and VBC \cite{fang2014automatic}. We also consider SPC, which is an advanced version of \cite{nishiyama2009sensation}, as described in \cite{yan2013learning}. Additionally, we adopt a recent aesthetics ranking method \cite{kong2016photo} combined with sliding window strategy as a baseline: ARC. We select the crop as the one with the highest ranking score from sliding windows. The comparison results on MSR-ICD and FLMS datasets are demonstrated in Table \ref{table3} and Table \ref{table4}, respectively. As seen, our cropping method achieves the best performance in both datasets. Qualitative results on MSR-ICD and FLMS datasets are presented in Fig. \ref{fig:fig10}.
\section{Conclusions}
\label{sec:conclusion}
In this work, we propose a deep learning based photo cropping approach, driven by human attention box prediction and aesthetics assessment. The proposed deep model is decomposed into two sub-networks: Attention Box Prediction (ABP) network and Aesthetics Assessment (AA) network, which share multiple convolution layers at the bottom. The proposed method approaches photo cropping in a determining-adjusting manner. It infers initial crop as a bounding box covering the visually important area (attention-aware determining), and then selects the best crop with highest aesthetic quality from a few cropping candidates generated around the initial crop (aesthetic-based adjusting). Our extensive experimental analyses demonstrate that our solution achieves superior performance in comparison to the state-of-the-art.

{\small
\bibliographystyle{ieee}
\bibliography{egbib}

\begin{thebibliography}{10}\itemsep=-1pt

\bibitem{borji2012boosting}
A.~Borji.
\newblock Boosting bottom-up and top-down visual features for saliency
  estimation.
\newblock In {\em CVPR}, 2012.

\bibitem{bruce2006saliency}
N.~Bruce and J.~Tsotsos.
\newblock Saliency based on information maximization.
\newblock {\em NIPS}, 2006.

\bibitem{Chen_2016_CVPR}
J.~Chen, G.~Bai, S.~Liang, and Z.~Li.
\newblock Automatic image cropping : A computational complexity study.
\newblock In {\em CVPR}, 2016.

\bibitem{datta2006studying}
R.~Datta, D.~Joshi, J.~Li, and J.~Z. Wang.
\newblock Studying aesthetics in photographic images using a computational
  approach.
\newblock In {\em ECCV}, 2006.

\bibitem{deng2016image}
Y.~Deng, C.~C. Loy, and X.~Tang.
\newblock Image aesthetic assessment: An experimental survey.
\newblock {\em arXiv preprint arXiv:1610.00838}, 2016.

\bibitem{dhar2011high}
S.~Dhar, V.~Ordonez, and T.~L. Berg.
\newblock High level describable attributes for predicting aesthetics and
  interestingness.
\newblock In {\em CVPR}, 2011.

\bibitem{everingham2010pascal}
M.~Everingham, L.~Van~Gool, C.~K. Williams, J.~Winn, and A.~Zisserman.
\newblock The {PASCAL Visual Object Classes (VOC)} challenge.
\newblock {\em IJCV}, 2010.

\bibitem{fang2014automatic}
C.~Fang, Z.~Lin, R.~M¡¦ech, and X.~Shen.
\newblock Automatic image cropping using visual composition, boundary
  simplicity and content preservation models.
\newblock In {\em ACMMM}, 2014.

\bibitem{girshickICCV15fastrcnn}
R.~Girshick.
\newblock Fast {R-CNN}.
\newblock In {\em ICCV}, 2015.

\bibitem{girshick2014rich}
R.~Girshick, J.~Donahue, T.~Darrell, and J.~Malik.
\newblock Rich feature hierarchies for accurate object detection and semantic
  segmentation.
\newblock In {\em CVPR}, 2014.

\bibitem{goferman2012context}
S.~Goferman, L.~Zelnik-Manor, and A.~Tal.
\newblock Context-aware saliency detection.
\newblock {\em IEEE PAMI}, 2012.

\bibitem{harel2006graph}
J.~Harel, C.~Koch, P.~Perona, et~al.
\newblock Graph-based visual saliency.
\newblock In {\em NIPS}, 2006.

\bibitem{he2014spatial}
K.~He, X.~Zhang, S.~Ren, and J.~Sun.
\newblock Spatial pyramid pooling in deep convolutional networks for visual
  recognition.
\newblock In {\em ECCV}, 2014.

\bibitem{hou2012image}
X.~Hou, J.~Harel, and C.~Koch.
\newblock Image signature: Highlighting sparse salient regions.
\newblock {\em IEEE PAMI}, 2012.

\bibitem{hou2009dynamic}
X.~Hou and L.~Zhang.
\newblock Dynamic visual attention: Searching for coding length increments.
\newblock In {\em NIPS}, 2009.

\bibitem{itti1998model}
L.~Itti, C.~Koch, and E.~Niebur.
\newblock A model of saliency-based visual attention for rapid scene analysis.
\newblock {\em IEEE PAMI}, 1998.

\bibitem{jia2014caffe}
Y.~Jia, E.~Shelhamer, J.~Donahue, S.~Karayev, J.~Long, R.~Girshick,
  S.~Guadarrama, and T.~Darrell.
\newblock Caffe: Convolutional architecture for fast feature embedding.
\newblock {\em arXiv preprint arXiv:1408.5093}, 2014.

\bibitem{jiang2015salicon}
M.~Jiang, S.~Huang, J.~Duan, and Q.~Zhao.
\newblock {SALICON}: Saliency in context.
\newblock In {\em CVPR}, 2015.

\bibitem{judd2009learning}
T.~Judd, K.~Ehinger, F.~Durand, and A.~Torralba.
\newblock Learning to predict where humans look.
\newblock In {\em ICCV}, 2009.

\bibitem{ke2006design}
Y.~Ke, X.~Tang, and F.~Jing.
\newblock The design of high-level features for photo quality assessment.
\newblock In {\em CVPR}, 2006.

\bibitem{kong2016photo}
S.~Kong, X.~Shen, Z.~Lin, R.~Mech, and C.~Fowlkes.
\newblock Photo aesthetics ranking network with attributes and content
  adaptation.
\newblock In {\em ECCV}, 2016.

\bibitem{li2014secrets}
Y.~Li, X.~Hou, C.~Koch, J.~M. Rehg, and A.~L. Yuille.
\newblock The secrets of salient object segmentation.
\newblock In {\em CVPR}, 2014.

\bibitem{liu2015predicting}
N.~Liu, J.~Han, D.~Zhang, S.~Wen, and T.~Liu.
\newblock Predicting eye fixations using convolutional neural networks.
\newblock In {\em CVPR}, 2015.

\bibitem{liu2016ssd}
W.~Liu, D.~Anguelov, D.~Erhan, C.~Szegedy, S.~Reed, C.-Y. Fu, and A.~C. Berg.
\newblock {SSD}: Single shot multibox detector.
\newblock In {\em ECCV}, 2016.

\bibitem{lu2014rapid}
X.~Lu, Z.~Lin, H.~Jin, J.~Yang, and J.~Z. Wang.
\newblock {RAPID}: Rating pictorial aesthetics using deep learning.
\newblock In {\em ACMMM}, 2014.

\bibitem{lu2015R}
X.~Lu, Z.~Lin, H.~Jin, J.~Yang, and J.~Z. Wang.
\newblock Rating image aesthetics using deep learning.
\newblock In {\em IEEE TMM}, 2015.

\bibitem{lu2015deep}
X.~Lu, Z.~Lin, X.~Shen, R.~Mech, and J.~Z. Wang.
\newblock Deep multi-patch aggregation network for image style, aesthetics, and
  quality estimation.
\newblock In {\em ICCV}, 2015.

\bibitem{mai2016composition}
L.~Mai, H.~Jin, and F.~Liu.
\newblock Composition-preserving deep photo aesthetics assessment.
\newblock In {\em CVPR}, 2016.

\bibitem{marchesotti2009framework}
L.~Marchesotti, C.~Cifarelli, and G.~Csurka.
\newblock A framework for visual saliency detection with applications to image
  thumbnailing.
\newblock In {\em ICCV}, 2009.

\bibitem{marchesotti2011assessing}
L.~Marchesotti, F.~Perronnin, D.~Larlus, and G.~Csurka.
\newblock Assessing the aesthetic quality of photographs using generic image
  descriptors.
\newblock In {\em ICCV}, 2011.

\bibitem{murray2012ava}
N.~Murray, L.~Marchesotti, and F.~Perronnin.
\newblock {AVA}: A large-scale database for aesthetic visual analysis.
\newblock In {\em CVPR}, 2012.

\bibitem{nishiyama2009sensation}
M.~Nishiyama, T.~Okabe, Y.~Sato, and I.~Sato.
\newblock Sensation-based photo cropping.
\newblock In {\em ACMMM}, 2009.

\bibitem{Pan_2016_CVPR}
J.~Pan, E.~Sayrol, X.~Giro-i Nieto, K.~McGuinness, and N.~E. O'Connor.
\newblock Shallow and deep convolutional networks for saliency prediction.
\newblock In {\em CVPR}, 2016.

\bibitem{park2012modeling}
J.~Park, J.-Y. Lee, Y.-W. Tai, and I.~S. Kweon.
\newblock Modeling photo composition and its application to photo
  re-arrangement.
\newblock In {\em ICIP}, 2012.

\bibitem{ren2015faster}
S.~Ren, K.~He, R.~Girshick, and J.~Sun.
\newblock Faster {R-CNN}: Towards real-time object detection with region
  proposal networks.
\newblock In {\em NIPS}, 2015.

\bibitem{russakovsky2015imagenet}
O.~Russakovsky, J.~Deng, H.~Su, J.~Krause, S.~Satheesh, S.~Ma, Z.~Huang,
  A.~Karpathy, A.~Khosla, M.~Bernstein, et~al.
\newblock Imagenet large scale visual recognition challenge.
\newblock {\em IJCV}, 2015.

\bibitem{simonyan2014very}
K.~Simonyan and A.~Zisserman.
\newblock Very deep convolutional networks for large-scale image recognition.
\newblock {\em arXiv preprint arXiv:1409.1556}, 2014.

\bibitem{su2011scenic}
H.-H. Su, T.-W. Chen, C.-C. Kao, W.~H. Hsu, and S.-Y. Chien.
\newblock Scenic photo quality assessment with bag of aesthetics-preserving
  features.
\newblock In {\em ACMMM}, 2011.

\bibitem{suh2003automatic}
B.~Suh, H.~Ling, B.~B. Bederson, and D.~W. Jacobs.
\newblock Automatic thumbnail cropping and its effectiveness.
\newblock In {\em ACM UIST}, 2003.

\bibitem{sun2013scale}
J.~Sun and H.~Ling.
\newblock Scale and object aware image thumbnailing.
\newblock {\em IJCV}, 2013.

\bibitem{tang2014blind}
H.~Tang, N.~Joshi, and A.~Kapoor.
\newblock Blind image quality assessment using semi-supervised rectifier
  networks.
\newblock In {\em CVPR}, 2014.

\bibitem{vig2014large}
E.~Vig, M.~Dorr, and D.~Cox.
\newblock Large-scale optimization of hierarchical features for saliency
  prediction in natural images.
\newblock In {\em CVPR}, 2014.

\bibitem{Wang2015}
W.~Wang, J.~Shen, and F.~Porikli.
\newblock Saliency-aware geodesic video object segmentation.
\newblock In {\em CVPR}, 2015.

\bibitem{wang2015consistent}
W.~Wang, J.~Shen, and L.~Shao.
\newblock Consistent video saliency using local gradient flow optimization and
  global refinement.
\newblock {\em IEEE TIP}, 2015.

\bibitem{wang2016correspondence}
W.~Wang, J.~Shen, L.~Shao, and F.~Porikli.
\newblock Correspondence driven saliency transfer.
\newblock {\em IEEE TIP}, 2016.

\bibitem{Wang2017}
W.~Wang, J.~Shen, R.~Yang, and F.~Porikli.
\newblock Saliency-aware video object segmentation.
\newblock {\em IEEE PAMI}, 2017.

\bibitem{wang2016stereoscopic}
W.~Wang, J.~Shen, Y.~Yu, and K.-L. Ma.
\newblock Stereoscopic thumbnail creation via efficient stereo saliency
  detection.
\newblock {\em IEEE TVCG}, 2016.

\bibitem{yan2013learning}
J.~Yan, S.~Lin, S.~Bing~Kang, and X.~Tang.
\newblock Learning the change for automatic image cropping.
\newblock In {\em CVPR}, 2013.

\bibitem{zhang2008sun}
L.~Zhang, M.~H. Tong, T.~K. Marks, H.~Shan, and G.~W. Cottrell.
\newblock {SUN}: A bayesian framework for saliency using natural statistics.
\newblock {\em Journal of vision}, 2008.

\end{thebibliography}
}

\end{document}